# A Masked Reverse Knowledge Distillation Method Incorporating Global and Local Information for Image Anomaly Detection

Yuxin Jiang[a], Yunkang Cao[a], Weiming Shen[a],*

[a]School of Mechanical Science and Engineering, Huazhong University of Science and Technology, Wuhan, 430000, Hubei, China

**Abstract**

Knowledge distillation is an effective image anomaly detection and localization scheme. However, a major drawback of this scheme is its tendency to overly generalize, primarily due to the similarities between input and supervisory signals. In order to address this issue, this paper introduces a novel technique called masked reverse knowledge distillation (MRKD). By employing image-level masking (ILM) and feature-level masking (FLM), MRKD transforms the task of image reconstruction into image restoration. Specifically, ILM helps to capture global information by differentiating input signals from supervisory signals. On the other hand, FLM incorporates synthetic feature-level anomalies to ensure that the learned representations contain sufficient local information. With these two strategies, MRKD is endowed with stronger image context capture capacity and is less likely to be overgeneralized. Experiments on the widely-used MVTec anomaly detection dataset demonstrate that MRKD achieves impressive performance: image-level 98.9% AU-ROC, pixel-level 98.4% AU-ROC, and 95.3% AU-PRO. In addition, extensive ablation experiments have validated the superiority of MRKD in mitigating the overgeneralization problem.

## 1. Introduction

Image anomaly detection (AD) aims to identify and localize heterogeneous regions in homogeneous images, which has been extensively applied in many fields, such as product defect detection [1], medical diagnosis [2], and video surveillance [3, 4]. In terms of industrial scenarios, such as surface defect detection, collecting abnormal samples is costly and time-consuming, which results in insufficient abnormal images and limits the practical applications of supervised AD methods [5]. To address this issue, a profusion of unsupervised methods has been introduced, where only normal samples are available in training.

Among existing unsupervised AD methods, knowledge distillation-based methods have attracted much popularity because of their superior detection performance. These methods usually contain a teacher-student pair, where a teacher network provides a supervisory signal to instruct a student network to reconstruct normal data. Their underlying assumption is that if the student network only perceives normal data, it will exhibit distinct behavior compared to the teacher network when presented with outliers. However, this core assumption may not always hold. In situations where the input samples deviate only slightly from normal data, the student network might unexpectedly generalize to anomalies and reconstruct abnormal representations well. This unforeseen generalization leads to a lack of apparent discrepancies between abnormal samples in the teacher-student pair, thereby compromising the performance of anomaly detection. In this paper, we refer to this unanticipated generalization phenomenon as overgeneralization.

To tackle the problem, researchers have proposed various solutions. One feasible solution is to incorporate memory modules and utilize the memorized normal prototypes to guide reconstruction [9, 10]. Another approach involves optimizing the discrepancy distributions of anomaly and anomaly-free samples collaboratively to boost the prediction reliability [22]. Besides, a new knowledge distillation-based AD method RD4AD introduces a reversed teacher-student model and a bottleneck module to enlarge the representation discrepancy of anomalies [26]. However, the performance of these methods remains limited due to the equivalence between input and supervisory signals. Concretely, as shown in Fig. 1 (a), normal features distilled by the teacher network are taken as both input and supervisory signals, which causes the student network to replicate inputs instead of deepening its understanding of image context [6].

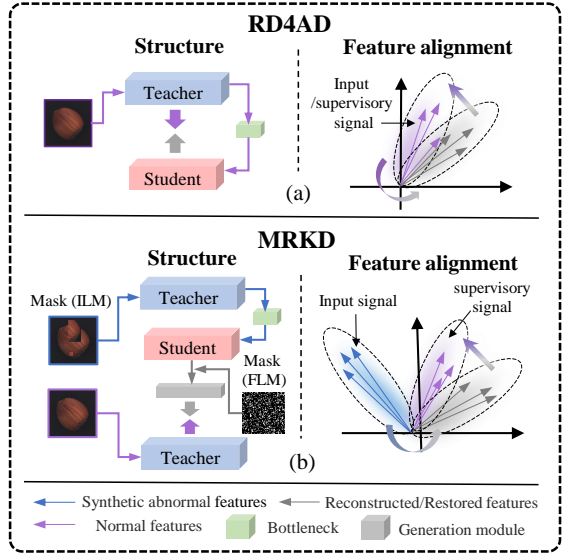

**Fig. 1.** Comparison between RD4AD and the proposed MRKD. RD4AD only perceives normal images, while MRKD is trained to restore the normal appearance of the synthetic abnormal features, learning representations biased towards image context.

To mitigate the aforementioned overgeneralization problem, this paper proposes a method called Masked Reverse Knowledge Distillation (MRKD) as shown in Fig. 1 (b). This method aims to translate the reconstruction task into a restoration task using two masking strategies: image-level masking (ILM) and feature-level masking (FLM). ILM is particularly important as it encourages the student network to learn representations that incorporate global information from large-scale image contexts. To achieve this, an anomaly-free image is first randomly masked with some image patches, and a teacher network is introduced to extract normal and abnormal features from the original and masked images, respectively. The student network is then trained to regress the normal features (the supervisory signal) with the abnormal features (the input signal), breaking the equivalence between the input and output. With ILM, the student network is forced to analyze the global information to conjecture the masked regions, promoting the learning of high semantic level representations. However, this restoration procedure emphasizes the holistic semantic but neglects small-scale local context and the correlation between adjacent pixels, i.e., local information, which may result in distortions of restored features. Therefore, the FLM strategy is proposed to compensate for the missing local information in ILM. In particular, feature-level anomalies are generated by masking random pixels of the features from the student network, and a simple block is used to restore the anomalies. Since the neighbor pixels already contain adequate information about the masked pixels, the student network can rely on local correlation to infer the missing areas. This procedure endows the student network with the capability to capture local information, which promotes finer-grained restoration. The two masking strategies complement each other from the image and feature levels to achieve the restoration of anomalies. As demonstrated in Fig. 2, the features of RD4AD still contain significant anomalies, while the proposed MRKD can effectively infer "normal-like" appearances of the abnormal regions, thereby mitigating the overgeneralization problem.

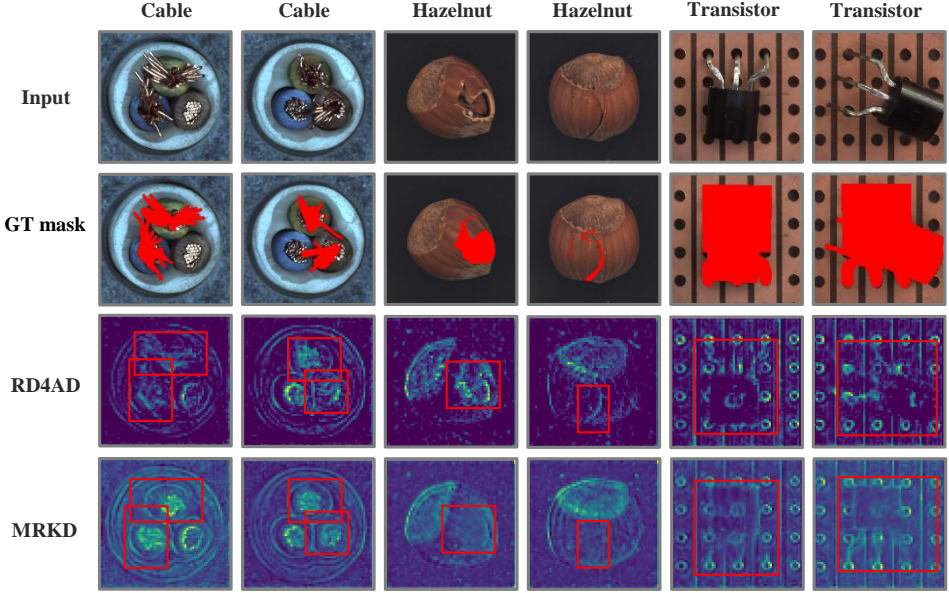

**Fig. 2.** RD4AD tends to overgeneralize to anomalies, while the proposed MRKD method successfully restores abnormally appearing regions back to a normal appearance. **Input:** Test images with defects. **GT mask:** Input images with ground-truth localization region masked in red. **RD4AD:** Reconstructed features generated by RD4AD. **MR**KD: Restored features generated by MRKD. Red rectangles are used to mark the restored abnormal regions.

The contributions of this paper are summarized as follows:
- The Masked Reverse Knowledge Distillation (MRKD) method is proposed to mitigate overgeneralization in image anomaly detection by transforming the pixel reconstruction task into restoration from abnormal to normal.
- The image-level masking strategy (ILM) is introduced to differentiate between input and supervisory signals using synthetic anomalies, enabling the student network to utilize global information for generating high-level semantic features.
- The feature-level masking strategy (FLM) is developed to facilitate the learning of representations with sufficient local information for fine-grained restoration by incorporating the restoration of feature-level anomalies into the student network optimization.

The remainder of this paper is organized as follows: Section 2 provides a comprehensive review of the related work. The proposed MRKD is illustrated in Section 3 in detail. Section 4 presents experiments on three public datasets and several ablation studies to verify the effectiveness of the proposed method. Section 5 concludes the paper and discusses some future work.

## 2. Related work

2.1 Reconstruction-based methods

At present, the reconstruction-based methods mainly utilize Autoencoder (AE) [7], Variational Autoencoders (VAE) [13], and Generative Adversarial Nets (GAN) [5, 14-15]. Since these methods are trained for reconstruction with only normal images, they will fail to accurately reconstruct abnormal areas never observed during training. In this case, the discrepancies between an input image and the corresponding reconstructed image could indicate anomaly degrees. However, in practical scenarios, these networks have powerful generalization abilities and may reconstruct even abnormal regions well [27]. Some methods have been developed for this issue. For instance, MemAE [9] proposes a memory-augmented autoencoder, which stores normal features as raw materials for reconstruction. MGNAD [10] further introduces two loss functions to enhance the compactness and separability of features stored in the memory module. RIAD [12] leverages iterative inpaintings for optimizing autoencoding to suppress the overgeneralization problem. DRAEM[16] combines the reconstruction network with a discrimination network to estimate an accurate decision boundary that separates anomalies from anomaly-free samples well. However, these methods may still suffer from limited reconstruction accuracies due to the deficiency of feature-level discriminative information.

Although the proposed MKD entails training the network to reconstruct anomalies to normal, it distinguishes itself from reconstruction-based methods in two key ways. First, the proposed method for detecting anomalies utilizes features extracted by a teacher network for supervisory signals, eschewing the use of generated segmentation masks or normal features from memory banks that are common in reconstruction-based methods. Second, unlike most reconstruction-based methods that require a generation network with encoder and decoder networks, along with a discrimination network or memory banks, the proposed method only requires the optimization of a student network and a simple generation module. This streamlined approach leads to a more efficient and concise structure for anomaly detection.

2.2 Embedding similarity-based methods

Embedding similarity-based methods have emerged as powerful techniques for detecting anomalies. These methods utilize pre-trained neural networks to extract distinctive features from intricate data structures, allowing them to identify anomalous instances based on their similarity values relative to normal data. By measuring the degree of similarity between extracted features and test data, these methods can effectively localize anomalous instances in various applications. For example, SPADE [18] detects abnormal areas by testing the pixel-level correspondence between the input images and nearest normal images, whereas its test complexity increases linearly with the increase of training data. To reduce calculation complexity, PaDiM [19] calculates the feature similarity based on the Mahalanobis distance after estimating the patch-level feature Gaussian distribution moments. Nonetheless, the efficacy of this method depends on image alignment. CPMF [20] and PatchCore [21] further propose to use a minimum selective feature set as a query set, which significantly improves the inference efficiency but still suffers in accurately utilizing both global and local information. Embedding-based methods excel in detecting anomalies for various data types, but there are still challenges in terms of accuracy and computational complexity.

## 2.3 Knowledge distillation-based methods

The knowledge distillation-based methods have been gaining popularity in recent years as a means to embed comprehensive feature information. This scheme contains a teacher-student pair where a knowledgeable teacher network acts as a pre-trained feature extractor and an awkward student network only trained to reconstruct normal data. Since the student network is not exposed to abnormal samples, its poor reconstruction ability for anomalies leads to significant discrepancies in teacher-student pairs, which can be leveraged to detect anomalies. In US [23], a teacher-student framework is utilized to differentiate anomalies by incorporating predictive uncertainty from a group of student networks and regression errors associated with a teacher network. STFPM [24] simplifies US [23] and utilizes a single teacher-student pair to obtain localize anomalies. Moreover, IKD [25] finds that STFPM [24] suffers from severe overfitting problems and proposes to distill informative knowledge in homogeneous normal images. RD4AD [26] trains a transposed distillation structure to avert the direct input of raw data and adds a bottleneck module for compressing redundant information, thereby increasing the discrepancy in teacher-student pairs. However, the student network in these approaches may even generalize to anomalies and reconstruct them well, leading to subpar detection performance.

Different from previous methods, MKRD explicitly trains the student networks to collaboratively restore the image and feature-level synthetic anomalies, enhancing the network's understanding of image context. By doing so, the method can better distinguish between normal and anomalous samples resulting in improved detection performance.

## 3. Proposed Masked Reverse Knowledge Distillation Method

### 3.1 Problem definition and method overview

**Problem Statement.** Given a training dataset $D_{train}$ containing a set of normal samples ($x_n \in D_{train}$) and a testing dataset $D_{test}$ consisting of both normal and abnormal samples ($x_t \in D_{test}$), our goal is to train a network to identify and further localize anomalies in $D_{test}$. Exiting methods have made tremendous efforts to constrain the awareness of the network under the normal pattern. However, the equivalent input and supervisory signals result in a poor capability to extract high-sematic features in most methods, thus, the student network can still generalize to anomalies.

**Overview of the proposed Method.** The proposed MRKD can restore abnormal features to normal ones. With MRKD, the restored "normal features" will considerably deviate from the original abnormal features, while the discrepancy can represent the anomaly localization result. As depicted in Fig. 3, MRKD is composed of image- and feature-level masking designs. Specifically, in ILM, we first apply a data augmentation strategy called NSA [28] to mask some patches of anomaly-free images and generate synthetic abnormal samples. These normal and synthetic abnormal images are then fed into a teacher network to extract their features, which are used as input and supervisory signals for the student network. The student network is trained to learn contextual relationships and restore abnormal features to normal features while preserving the normal regions. Subsequently, in FLM, we randomly mask some pixel features from the student network and restore these synthetic feature-level anomalies through a simple generation module. The masked pixels can be extrapolated from visible neighboring patches, thus improving the utilization of local information. In the rest of this section, ILM and FLM are first illustrated in detail. Then, we introduce the testing phase, specifying the procedure to detect and localize anomalies. Finally, two masking strategies are analyzed from the aspects of form and function to emphasize their complementarity.

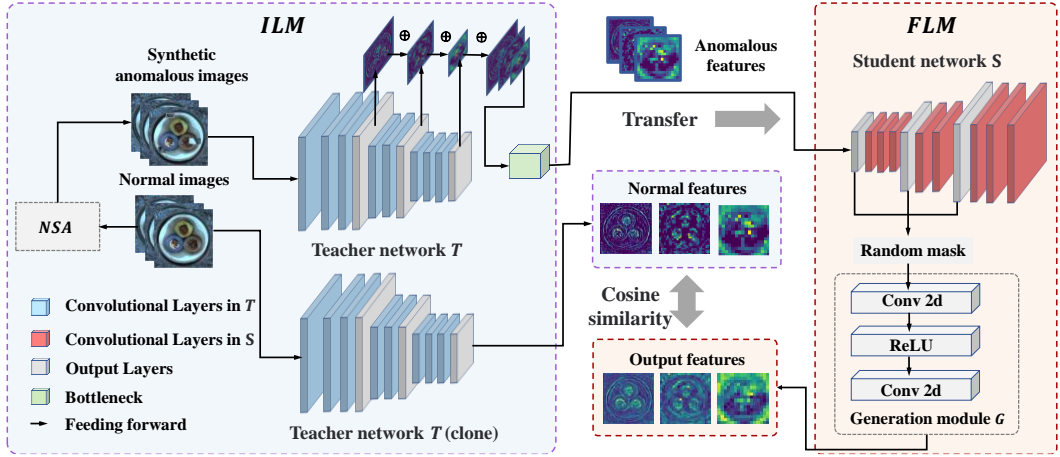

**Fig. 3.** Overview of the proposed MRKD method. In ILM, is utilized to synthesize abnormal images. Subsequently, the teacher network extracts features from abnormal and normal images, respectively. In FLM, output features of the student network are randomly masked, following which the generation module is employed to restore the masked features. Finally, the cosine similarity function is used to match the outputs with anomaly-free features, forcing the restored features to approximate the normal manifold. It is noteworthy that during inference, NSA is not performed on test images.

### 3.2 ILM

**Masking.** To endow the student network with the ability to detect and restore real anomalies into normal features, we need to generate synthetic anomalies similar to real ones and exploit them as training data. We follow NSA [28] to mask some information in $x_n$. Specifically, some patches are cropped from an image in $D_{train}$ and seamlessly bent into $x_n$ to mask some normal information, getting a synthetic abnormal image $x_a$. These abnormal patches are parts of normal images with semantically plausible non-anomaly content, thus highlighting the higher requirement for semantic understanding of the network. Moreover, these patches are replicated in foreground regions, preventing the network from focusing on the background.

**Teacher network.** We use a network WideResNet50 [29] pre-trained on ImageNet [30] as the teacher network $T$ to extract comprehensive semantic information. $T$ is parameterized by $\theta_T$ that is frozen during the training stage. The features extracted by $T$ are defined as $f_T^l$, $l \in \{1,2,\cdots N\}$, where $l$ represents the $l$-th block in $T$. $x_n$ and $x_a$ are input into $T$ to obtain normal and abnormal features taken as input signals and supervisory signals, respectively. Then, only abnormal features are fed into a bottleneck module to eliminate redundancy, obtaining rich yet compact features:

$$\begin{cases} f_{T,n}^l = T(x_n) \\ f_{T,a}^l = T(x_a) \\ f_B = B(f_{T,a}^l) \end{cases} \quad (1)$$

where, $f_{T,n}^l$ and $f_{T,a}^l$ are normal and abnormal features, respectively, $B$ denotes bottleneck module with random weights $\theta_B$, and $f_B$ is the compact abnormal feature condensed by $B$.

**Principle Discussion.** ILM breaks the equivalence between input and supervisory signals, which converts the pixel restoration to the restoration of anomalies. It is necessary to implement this transformation due to the little understanding of semantics in traditional methods. Specifically, these methods set equivalent input and supervisory signals while the convolutional neural network (CNN) architecture is inclined to encode much low-scale textural information. Therefore, the student network tends to learn a low-level pixel-to-pixel replication during training. However, since the textures of low-scale abnormal regions are markedly similar to those of normal samples, these methods easily suffer from overgeneralizing to abnormal features. Therefore, we use the distilled normal feature restores as supervisory signals and abnormal features as input signals in order to bias the network to focus on large-scale global information.

Fig. 4 further exemplifies the definition of global information and its contribution to achieving anomaly detection. To restore some pixels (masked with red) in the abnormal regions (marked with grey rectangles), the student network tends to rely on the information contained in nearby pixels (masked with green rectangles) rather than holistic semantics. However, the green rectangle only contains abnormal features that help little in restoring "normal-like" features. Under this setting, the student network can only give up relying on low-scale features to regress supervisory signals and pay more attention to the broad contextual relationship, i.e., global information, to extrapolate the "normal-like" appearance. The leverage of global information benefits the student network by deepening a comprehensive understanding of semantics rather than simply memorizing every pixel detail of input features for restoration purposes.

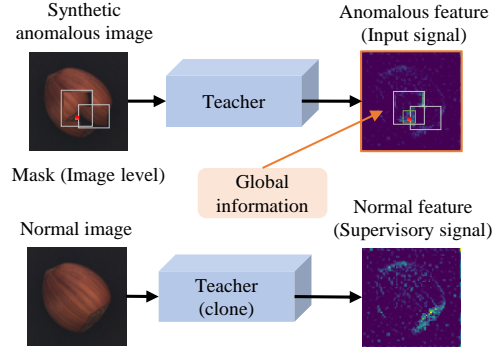

**Fig. 4.** The learning of global information. The input and supervisory signals are the distilled feature maps from the teacher network. Global information refers to the contextual relationship in the abnormal feature. It serves as a critical clue to conjecture the normal appearance of the anomalies marked with grey rectangles.

### 3.3 FLM

**Student network.** The student network denoted as $S$ has a symmetrical but reversed structure to $T$ and is randomly initialized by parameter $\theta_S$. After the extraction of $f_{T,n}^l$ and $f_B$, we only input $f_B$ into $S$, yielding features $f_S^l = S(f_B)$. It is critical to optimize the performance of $S$ to accomplish feature restoration. The masking in ILM encourages $S$ to focus on global information, while the masking in FLM described next emphasizes the importance of local information.

**Masking.** Feature-level anomaly is synthesized following Masked Generative Distillation (MGD) [31], i.e., randomly masking all areas of $f_S^l$, whether abnormal or normal. The masking ratio is set to low (see Fig. 5). This is because a high masking ratio will lead to a large amount of information loss, thus degrading the restoration accuracy. Besides, the restoration of highly sparse features enhances the understanding of the global relationship against our purpose to optimize the capability to extract local information. Then, a generation module $G$ is employed to restore the masked feature as the full normal feature $f_{T,n}^l$ extracted by the teacher network. The process can be formulated as:

$$M^l(h,w) = \begin{cases} 0, & if\ R^l(h,w) < \lambda \\ 1, & otherwise \end{cases} \quad (2)$$

$$f_G^l = G(f_S^l \odot M^l) = W_{l2}(ReLU(W_{l1}(f_S^l \odot M^l))) \quad (3)$$

where, $R^l(h,w)$ is a random number in (0, 1) at image coordinates $(h,w)$, and $M^l$ is the generated mask, $\lambda$ represents the masking ratio, $\odot$ operates the element-wise multiplication. $G$ denotes a generation module parameterized by $\theta_G$, which includes two convolution layers $W_{l1}$ and $W_{l2}$, one activation layer $ReLU$, and $f_G^l$ is the final restored feature by $G$.

**Feature alignment:** To restore features $f_G^l$ belonging to normal data manifold, we align them with $f_{T,n}^l$. Specifically, a vectorization function, denoted as $vec(\cdot)$, is introduced to convert $f_{T,n}^l$ and $f_G^l$ into 1-D vector. Then, the similarity between $f_{T,n}^l$ and $f_G^l$ is quantified using the cosine similarity function. The cosine similarity values of each layer are summed together to form the loss function, which is formulated as follows:

$$L = \sum_{l=1}^{N}(1 - \frac{vec(f_{T,n}^l)^T \cdot vec(f_G^l)}{\|vec(f_{T,n}^l)\|\|vec(f_G^l)\|}) \quad (4)$$

**Principle Discussion.** FLM aims to enhance the ability of the student network to acquire feature representations containing local information. This is achieved by restoring feature-level anomalies, as shown in Fig. 5. The restoration of missing pixels (marked with grey rectangles) is more determined by the remaining surrounding pixels because these surrounding pixels belong to embeddings of deeper layers already containing some information about the masked ones. Therefore, using partial pixels to restore complete features can strengthen the extraction of local information. Unlike global information which emphasizes holistic semantic information extraction and tends to overlook details when restoring features, FLM leverages local information to refine restored features. This ensures that the details of the features are not distorted even when generating "normal-like" features. FLM complements global information-based restoration by involving more pixels in the restoration process. This approach improves the representation power of local information in the used pixels, ultimately promoting the precision of the restored features.

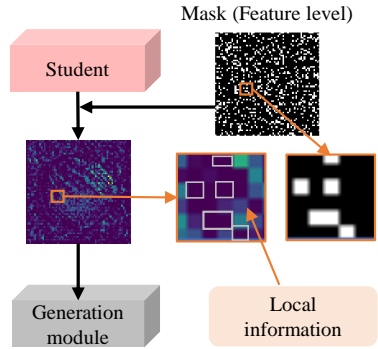

**Fig. 5.** The learning of local information. Local information refers to pixel correlation belonging to the patch in the orange rectangle. The student network can learn this information by inferring the lost pixels (feature-level anomalies) marked with a grey rectangle.

3.4 Inference

Recall that after training, the student network is endowed with the capability to restore abnormal features while maintaining the normal features unchanged. Therefore, the discrepancies between features $f_T^l$ and $f_G^l$ extracted by the teacher and student networks can provide a compelling clue to localize anomalies. To develop a pixel-level anomaly score map for anomaly localization, we first calculate cosine similarity between $f_T^l$ and $f_G^l$ along the channel axis and then up-sample the cosine similarity map to image size. Moreover, many researchers have proved that the anomaly score map obtained by gathering multi-scale features together contains both low-scale information (i.e., texture and edge) and large-scale information (i.e., structure and orientation), performing subpar anomaly localization [25-26]. Thus, the anomaly score map is defined as:

$$S_{AL}(h,w) = \sum_{l=1}^{N} \Phi(1 - \frac{\left(f_T^l(h,w)\right)^T \cdot f_G^l(h,w)}{\|f_T^l(h,w)\|\|f_G^l(h,w)\|}) \tag{5}$$

where, $S_{AL}(h,w)$ represents the anomaly score map of size $h \times w$, the $\Phi$ function performs a bilinear up-sampling operation.

For anomaly detection, the maximum value in $S_{AL}(h,w)$ can portray the biggest margin between $f_G^l$ and $f_{T,t}^l$. Hence, $S_{AD}$ is achieved by:

$$S_{AD} = \max(S_{AL}(h,w)) \tag{6}$$

To summarize, in the training stage, masking strategies are implemented at both the image and feature levels. In the testing stage, it is crucial to also carry out masking at the feature level to ensure that the testing samples align with the same domain as the training samples. Following this step, multi-scale cosine similarity maps are collected to derive anomaly scores.

The whole process of the proposed method is summarized in Algorithm 1.

---

**Algorithm 1:** Masked Reverse Knowledge Distillation

**Training stage:**

**Input:** $D_{train}$ of AD dataset, teacher network $T$, student network $S$, bottleneck module $B$, generation module $G$, hyper-parameter $\lambda$, and their parameters $\{\theta_T, \theta_S, \theta_B, \theta_G\}$

**Output:** The parameters $\{\theta_S, \theta_B, \theta_G\}$

/* Perform both ILM and FLM */

1:    Initialize $\theta_T$ with pre-trained weights
2:    Initialize $\{\theta_S, \theta_B, \theta_G\}$ with the random weights
3:    **for** iteration number **do**
4:       Randomly sample a batch of normal samples
5:       Construct synthetic abnormal samples by applying the NSA strategy
6:       Generate features $f_{T,n}^l$ and $f_B$ by $T$ and $B$ using Eq. (1)
7:       Generate features $f_s^l$ by $S$
8:       Mask some pixels of $f_s^l$ by $S$ and restore $f_G^l$ by $G$ using Eq. (2) and (3)
9:       Calculate the restoration loss using Eq. (4)
10:     Update $\{\theta_S, \theta_B, \theta_G\}$ using Stochastic Gradient
      **end for**

**Testing stage:**

**Input:** $D_{test}$ of AD dataset, the parameters $\{\theta_S, \theta_B, \theta_G\}$ with saved weights

**Output:** Anomaly scores $S_{AL}$ and $S_{AD}$

/* Perform only FLM */

1:    Follow the steps 6-8 of the training stage
2:    Calculate the anomaly scores $S_{AL}$ and $S_{AD}$ using Eq. (5) and (6)

3.5 Complementarity between two masking strategies

To further elaborate the principle of the proposed MRKD method, we analyze the respective properties of the proposed two masking strategies and verify their complementarity of anomaly detection and localization.

**From the perspective of form.** There are two major differences between ILM and FLM. Firstly, outlier information is inputted during the synthesis of abnormal samples in ILM, which destroys the holistic structure and context of the images, while masked features in FLM only lose parts of information without the addition of extra information. Secondly, the added outlier information in ILM possesses semantic plausibility, while the missing information in FLM does not make up a semantic segment.

Furthermore, it is worth noting that the proposed method does not necessitate synthetic anomalies that closely resemble real anomalies at both image and feature levels. This is due to the fact that the MRKD utilizes partial normal information to derive the complete normal appearance, instead of learning discriminative features between abnormal and normal instances. As a result, our method has lower demands for the similarity between synthetic and actual anomalies compared to other pseudo-anomalies augmentation methods such as NSA and CutPaste.

**From the perspective of function.** ILM is employed to destroy the equivalence of input and supervisory signals, which offers an opportunity to create a student network capable of extrapolating a "normal-like" appearance from anomalies. In addition, since the abnormal areas to be speculated are semantic entities with normal contents, contextual relationship plays an important role in restoration, thus encouraging the student network to capture global information. In FLM, missing pixels can be recovered from neighboring pixels, which enables the used pixels to contain more local information for refining the restored features. Therefore, FLM and ILM complement each other to mitigate the overgeneralization problem.

## 4. Experimental Validation and Evaluation

4.1 Datasets and Evaluation Metrics

**Datasets.** We evaluate the anomaly detection performance of the proposed method on datasets of MVTec AD [32], Magnetic Tile Defect (MTD) [33], and beanTech Anomaly Detection (BTAD) [34]. MVTec AD is a widely used dataset for industrial inspection. It has a total of 5354 high-resolution images, including five texture classes and ten object classes. Each sub-dataset is divided into a training set with only anomaly-free samples and a test set with both normal samples and various abnormal samples labeled with pixel-precise annotations. MTD dataset contains 925 defect-free and 392 abnormal magnetic tile images with different sizes. We comply with the setting in [35]: 80% of normal images are randomly selected for training, and the rest, along with all the abnormal images for testing. BTAD contains 2830 images in three categories composed of normal and abnormal images. All images in the three datasets are resized to 256 × 256 and normalized by the mean and variance of the ImageNet dataset for training and testing.

**Evaluation Metrics.** For evaluation, the standard metric Area Under the Receiver Operating Characteristics (AU-ROC) is adopted to measure the performance. Image-level AU-ROC (AU-ROC$_{IL}$) is used for anomaly detection, and pixel-level AU-ROC(AU-ROC$_{PL}$) is utilized for anomaly localization. The false-positive rate is controlled by a very high number of non-abnormal pixels when the abnormal area is tiny. Consequently, the value of AU-ROC$_{PL}$ is still low despite the existence of false-positive tests. Therefore, AU-ROC$_{PL}$ is biased in favor of anomalies occupying large areas. As a supplement, we also use the pre-region-overlap (AU-PRO) score as the evaluation index, which can equally weigh ground-truth abnormal areas regardless of the area size. Hence, the large anomaly bias will be penalized [21].

**Implementation Details.** We take WideResNet50 without the classification layers as the teacher network, which follows a bottleneck module set as [26] to reduce redundancy. The student network has a symmetrical but reversed structure to the teacher network, and the generating module includes two convolution layers and one ReLU activation layer. We train the models on batches of size 16 using an Adam optimizer adopted with a learning rate of 0.005 over 200 epochs. The hyperparameter $\lambda$ in Eq. (2) is fixed to 0.2. We use PyTorch V1.12.0 and train each model on a Nvidia GeForce RTX 3090 Ti GPU and an Intel i9@3.00GHz CPU.

4.2 Anomaly detection and localization

In Table 1, we benchmark the proposed method against nine state-of-the-art approaches to anomaly detection. The proposed MRKD achieves 98.9% AU-ROC, significantly outperforming all counterparts except PatchCore. Moreover, for cable and transistor categories where RD4AD performs weakly, the proposed method significantly outclasses RD4AD by 3.4% and 3.3% AU-ROC. To conceptualize the AD capability, we further visualize the anomaly scores of all test samples in Fig. 6. There is rarely an overlap of normal and abnormal distribution for all categories, which proves the strong discriminability of the proposed MRKD on anomalies.

Table 2 shows the anomaly localization performance. The results demonstrate that the average value of the proposed method significantly surpasses other methods, with the highest 98.4% AU-ROC. Besides, the AU-ROCs of 13 categories exceed 97.5%, indicating a pleasant adaption ability to various anomalies. In addition, we further assess the localization ability in terms of PRO. The proposed method also obtains the SOTA result, showing its performance improvement to RD4AD. Especially for the transistor class, the proposed method can surpass all compared methods by a considerable margin, notably over RD4AD by 13.4%.

The proposed approach was also assessed on the MTD dataset, which differs from the MVTec AD dataset as it consists of magnetic tile images of varying sizes. The results indicate that the proposed approach achieved excellent accuracy, while some spatially rigid methods such as PaDiM struggled to handle this type of problem. Table 3 shows that MRKD can achieve 98.7% AU-ROC over DifferNet by 10%. Additionally, the performance of MRKD was evaluated on the BTAD dataset and compared to other methods in Table 4. The results show that MRKD reached a new state-of-the-art of 98.0% AU-ROC, indicating strong anomaly detection capability.

**Table 1**

Anomaly detection performance (AU-ROC$_{IL}$) of different methods on the MVTec AD dataset. The best results are in bold, and values in parentheses represent changes relative to RD4AD, with "+" indicating an increase and "-" representing a decrease.

| Category | DRAEM [16] | RIAD [12] | Intra [17] | PatchCore [21] | P-SVDD [36] | SPADE [18] | PaDiM [19] | CutPaste [37] | STFPM [24] | NSA [28] | RD4AD [26] | MRKD |
|---|---|---|---|---|---|---|---|---|---|---|---|---|
| Carpet | 97 | 84.2 | 98.8 | **100** | 92.9 | - | 99.8 | 93.9 | - | 95.6 | 98.9 | **100**(+1.1) |
| Grid | 99.9 | 99.6 | 100.0 | 99.5 | 94.6 | - | 96.7 | **100** | - | 99.9 | **100** | **100**(+0.0) |
| Leather | **100** | **100** | 100.0 | 98.1 | 90.9 | - | **100** | **100** | - | 99.9 | **100** | **100**(+0.0) |
| Tile | 99.6 | 98.7 | 98.2 | 98.7 | 97.8 | - | 98.1 | 94.6 | - | **100** | 99.3 | 99.2(-0.1) |
| Wood | 99.1 | 93 | 97.5 | 98.2 | 96.5 | - | 99.2 | 99.1 | - | 97.5 | 99.2 | **99.4**(+0.2) |
| Bottle | 99.2 | 99.9 | **100** | **100** | 98.6 | - | 99.9 | 98.2 | - | 97.7 | **100** | **100**(+0.0) |
| Cable | 91.8 | 81.9 | 70.3 | **100** | 90.3 | - | 92.7 | 81.2 | - | 94.5 | 95 | 98.4(+3.4) |
| Capsule | 98.5 | 88.4 | 86.5 | **100** | 76.7 | - | 91.3 | **98.2** | - | 95.2 | 96.3 | 96.5(+0.2) |
| Hazelnut | **100** | 83.3 | 95.7 | 96.6 | 92 | - | 92 | 98.3 | - | 94.7 | 99.9 | 99.2(-0.7) |
| Metal nut | 98.7 | 88.5 | 96.9 | 98.1 | 94 | - | 98.7 | 99.9 | - | 98.7 | **100** | 100.0(+0.0) |
| Pill | 98.9 | 83.8 | 90.2 | 98.7 | 86.1 | - | 93.3 | 94.9 | - | **99.2** | 96.6 | 98.1(+1.5) |
| Screw | 93.9 | 84.5 | 95.7 | **100** | 81.3 | - | 85.8 | 88.7 | - | 90.2 | 97 | 98.2(+1.2) |
| Toothbrush | **100** | **100** | **100** | **100** | **100** | - | 96.1 | 99.4 | - | **100** | 99.5 | 97.8(-1.7) |
| Transistor | 93.1 | 90.9 | 95.8 | 99.2 | 91.5 | - | 97.4 | 96.1 | - | 95.1 | 96.7 | **100**(+3.3) |
| Zipper | **100** | 98.1 | 99.4 | 99.4 | 97.9 | - | 90.3 | 99.9 | - | 99.8 | 98.5 | 97.3(-1.2) |
| Average | 98.0 | 91.7 | 95 | **99.1** | 92.1 | 85.5 | 95.5 | 96.1 | 95.5 | 97.2 | 98.5 | 98.9(+0.4) |

## Table 2

Anomaly localization performance (AU-ROC$_{PL}$ / AU-PRO) on the MVTec AD dataset. The best results are in bold, and values in parentheses represent changes relative to RD4AD, with "+" indicating an increase and "-" representing a decrease.

| Category | DRAEM [16] | RIAD [12] | Intra [17] | PatchCore [21] | P-SVDD [36] | SPADE [18] | PaDiM [19] | CutPaste [37] | STFPM [24] | NSA [28] | RD4AD [26] | MRKD |
|---|---|---|---|---|---|---|---|---|---|---|---|---|
| Carpet | 95.5/- | 96.3/- | 99.2/- | 99/96.6 | 92.6/- | 97.5/94.7 | 99.1/96.2 | 98.3/- | 98.8/95.8 | 95.5/- | 98.9/97 | **99.3**(+0.4)/**98.1**(+1.1) |
| Grid | **99.7**/- | 98.8/- | 98.8/- | 98.7/96.0 | 96.2/- | 93.7/86.7 | 97.3/94.6 | 97.5/- | 99/96.6 | 99.2/- | 99.3/**97.6** | 99.3(+0)/97.5(-0.1) |
| Leather | 98.6/- | 99.4/- | 99.5/- | 99.3/98.9 | 97.4/- | 97.6/97.2 | 99.2/97.8 | **99.5**/- | 99.3/98 | **99.5**/- | 99.4/99.1 | **99.5**(+0.1)/**99.2**(+0.1) |
| Tile | 99.2/- | 89.1/- | 94.4/- | 98.7/87.3 | 91.4/- | 87.4/75.9 | 94.1/86 | 90.5/- | 97.4/92.1 | **99.3**/- | 95.6/90.6 | 95.9(+0.3)/91.9(+1.3) |
| Wood | 96.4/- | 85.8/- | 88.7/- | 98.2/89.4 | 90.8/- | 88.5/87.4 | 94.9/91.1 | 95.5/- | **97.2/93.6** | 90.7/- | 95.3/90.9 | 95.9(+0.6)/92.5(+1.6) |
| Bottle | **99.1**/- | 98.4/- | 97.1/- | 98.6/96.2 | 98.1/- | 98.4/95.5 | 98.3/94.8 | 97.6/- | 98.8/95.1 | 98.3/- | 98.7/96.6 | 98.7(+0)/**96.9**(+0.3) |
| Cable | 94.7/- | 84.2/- | 91/- | 98.4/92.5 | 96.8/- | 97.2/90.9 | 96.7/88.8 | 90/- | 95.5/87.8 | 96/- | 97.4/91 | **98.3**(+0.9)/**94.2**(+3.2) |
| Capsule | 95.5/- | 92.8/- | 97.7/- | 98.8/95.5 | 95.8/- | **99.0**/93.7 | 98.5/93.5 | 97.4/- | 98.3/92.2 | 97.6/- | 98.7/95.8 | 98.8(+0.1)/95.3(-0.5) |
| Hazelnut | **99.7**/- | 96.1/- | 98.3/- | 98.7/93.8 | 97.5/- | 99.1/95.4 | 98.2/92.6 | 97.3/- | 98.5/94.3 | 97.6/- | 98.9/**95.5** | 99.1(+0.2)/**95.5**(+0) |
| Metal nut | **99.5**/- | 92.5/- | 93.3/- | 98.4/91.4 | 98/- | 98.1/94.4 | 97.2/85.6 | 93.1/- | 97.6/**94.5** | 98.4/- | 97.3/92.3 | 97.6(+0.3)/91.7(-0.6) |
| Pill | 97.6/- | 95.7/- | 98.3/- | 97.4/93.2 | 95.1/- | 96.5/94.6 | 95.7/92.7 | 95.7/- | 97.8/96.5 | **98.5**/- | 98.2/96.4 | 98.5(+0.3)/**97.1**(+0.7) |
| Screw | 97.6/- | 98.8/- | 99.5/- | 99.4/97.9 | 95.7/- | 98.9/96 | 98.5/94.4 | 96.7/- | 98.3/93 | 96.5/- | **99.6**/98.2 | **99.6**(+0)/**98.4**(+0.2) |
| Toothbrush | 98.1/- | 98.9/- | 98.9/- | 98.7/91.5 | 98.1/- | 97,9/93.5 | 98.8/93.1 | 98.1/- | 98.9/92.2 | 94.9/- | **99.1/94.5** | **99.1**(+0)/94.1(-0.4) |
| Transistor | 90.9/- | 87.7/- | 96.1/- | 96.3/83.7 | 97/- | 94.1/87.4 | 97.5/84.5 | 93/- | 82.5/69.5 | 88/- | 92.5/78 | **98.1**(+5.6)/**91.4**(+13.4) |
| Zipper | 98.8/- | 97.8/- | 99.2/- | 98.8/91.7 | 95.1/- | 96.5/92.6 | 98.5/**95.9** | **99.3**/- | 98.5/95.2 | 94.2/- | 98.2/95.4 | 98.3(+0.1)/95.4(+0) |
| **Average** | 97.3/- | 94.2/- | 96.6/- | 98.1/93.4 | 95.7/- | 96.5/91.7 | 97.5/92.1 | 96/- | 96.5/92.1 | 96.3/- | 97.8/93.9 | **98.4**(+0.6)/**95.3**(+1.4) |

## Table 3

Anomaly detection performance (AU-ROC$_{IL}$) on the MTD dataset. The best results are in bold.

| MTD | GeoTrans[35] | OCSVM[35] | DSEBN[35] | GANomaly[35] | I-NN[35] | DifferNet[35] | Ours |
|---|---|---|---|---|---|---|---|
| AU-ROC(%) | 75.5 | 58.7 | 57.2 | 76.6 | 80.0 | 97.7 | **98.7** |

## Table 4

Anomaly localization performance (AU-ROC$_{PL}$) on the BTAD dataset. The best results are in bold.

| Categories | AE MSE [34] | AE MSE+SSIM [34] | VT-ADL [34] | MRKD |
|---|---|---|---|---|
| 0 | 0.49 | 0.53 | 0.99 | **0.97** |
| 1 | 0.92 | 0.96 | 0.94 | **0.97** |
| 2 | 0.95 | 0.89 | 0.77 | **1.00** |
| Average | 0.78 | 0.79 | 0.90 | **0.98** |

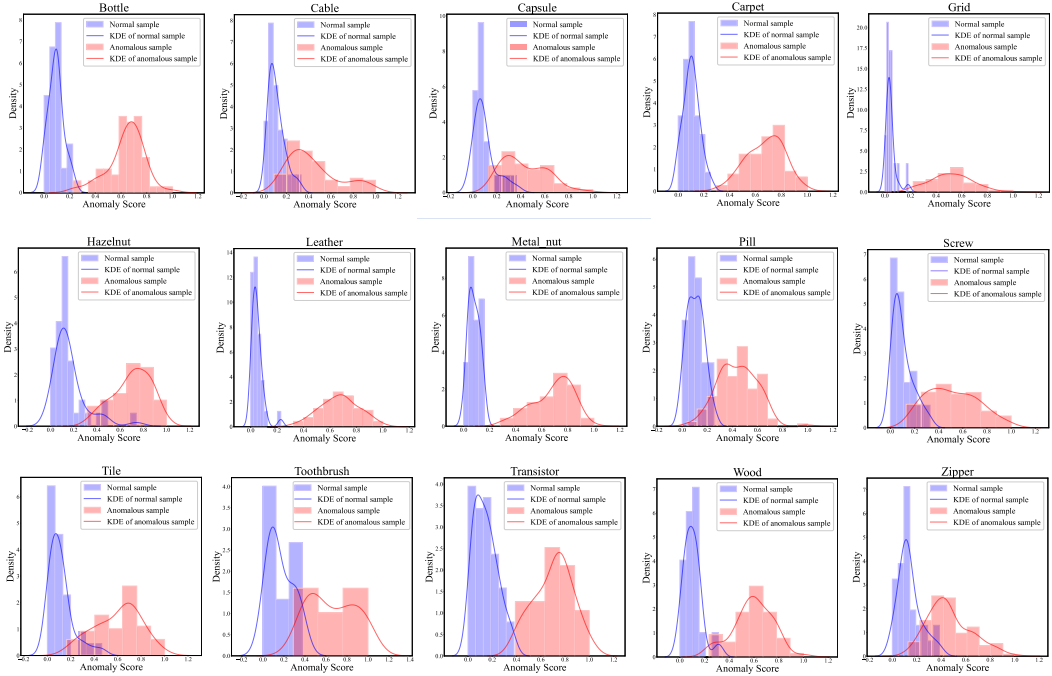

**Fig. 6.** Visualization of the distributions of anomaly scores for fifteen categories of MVTec. The kernel density estimation (KDE) is used to estimate probability densities of anomaly scores, which are then displayed as red (abnormal samples) and blue (normal samples) lines. The horizontal coordinate of the histogram represents anomaly scores from 0 to 1, and the height of the bars shows a density rather than a count.

4.3 Qualitative Results

The visualization of qualitative comparisons between the MRKD and RD4AD is presented. As shown in Fig. 7, RD4AD misses out on some flaws that are close to the normal distribution (such as Hazelnut Tile and Bottle categories), and it also struggles to identify misplaced anomaly-free patches (such as Cable and Transistor). In addition, RD4AD is sensitive to noises in images (as seen in the Zipper, Pill, and Carpet categories), which may lead to misdiagnosing normal regions as abnormal. Note that the proposed MRKD method considerably gains the accuracy of anomaly localization. For instance, it can localize both misplaced regions and original regions on the transistor dataset. Besides, MRKD exhibits excellent anti-noise capability, such as dealing with the irregular texture of the carpet and zipper or the disorderly red spots on pills, which demonstrates the superior performance of anomaly localization.

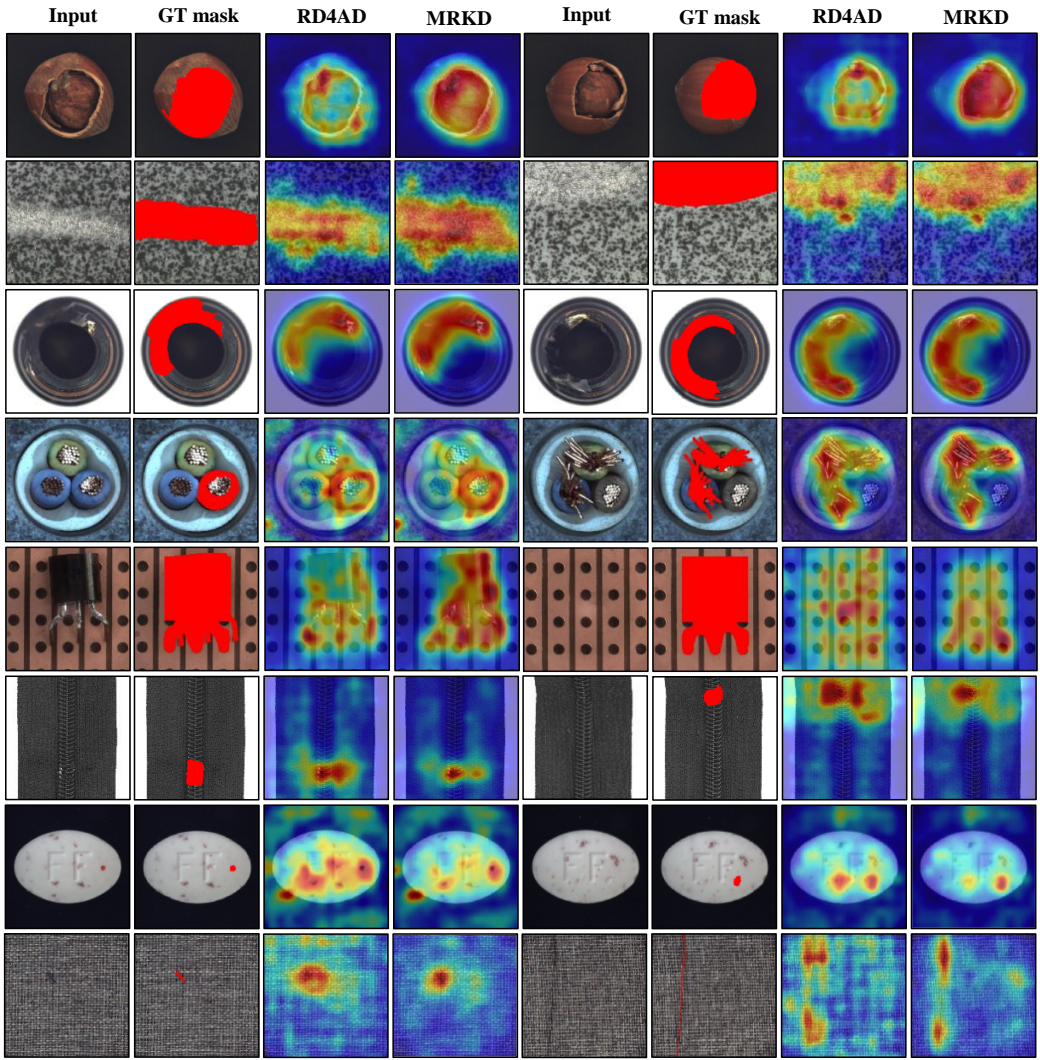

**Fig. 7:** Anomaly localization results of the MVTec AD datasets. **Input:** Test images with defects. **GT mask:** Input images with ground-truth localization regions masked in red. **RD4AD:** Anomaly localization heatmaps of RD4AD. **MRKD:** Anomaly localization heatmaps of MRKD. From top row to bottom row are Hazelnut, Tile, Bottle, Cable, Transistor, Zipper, Pill and Carpet categories.

4.4 Ablation Study

To gain a more profound understanding of the proposed approach, we conducted several ablation studies. We take RD4AD as the baseline and report the effectiveness of ILM and FLM in Table 5. Additionally, we explore how the two masking strategies complement each other and investigate the impact of hyperparameters.

**Table 5**

Ablation study of ILM and FLM. The best results are in bold.

| Baseline | ILM | FLM | AU-ROC$_{AD}$ | AU-ROC$_{AI}$ | AU-PRO |
|---|---|---|---|---|---|
| √ | | | 98.5 | 97.8 | 93.9 |
| √ | √ | | 98.7 | 98.2 | 95 |
| √ | | √ | **98.9** | 98 | 94.4 |
| √ | √ | √ | **98.9** | **98.4** | **95.3** |

**Contribution of ILM.** ILM is designed to encourage the student network to capture high-level semantics based on global correlation, which aids in restoring normal appearance from anomalies. To validate its efficiency, we visualize the features yielded by RD4AD and the proposed MRKD. As shown in Fig. 8, RD4AD reproduces the anomalies obviously, while MRKD can successfully restore abnormal regions to normal, generating glaring discrepancies to input features. Besides, the network consisting of only ILM *without FLM* can also produce "normal-like" features. Therefore, it can be inferred that the adoption of ILM boosts the representative capacity of global information, which facilitates recovering anomaly-free features from anomalies.

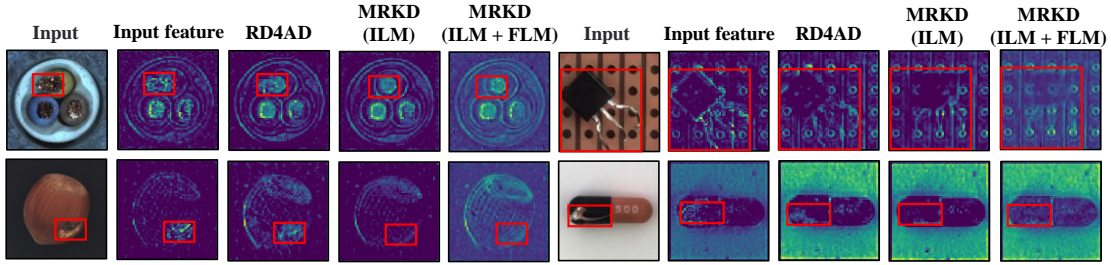

**Fig. 8.** Visualization of the features from the first output layer of RD4AD and the proposed MRKD. **Input:** Test images with defects. **Input feature:** Input features of the student network. **RD4AD:** Reconstructed features generated by the student network in RD4AD. **MRKD (ILM):** Restored features generated by the student network in MRKD considering only ILM. **MRKD (ILM + FLM):** Restored features generated by the student network in MRKD considering both ILM and FLM. Red boxes are used to highlight the abnormal areas and the restored "normal-like" areas.

**Contribution of FLM.** Recall that ILM promotes the use of global information. However, omitting local information will lead to inaccurate restoration that presumably results in localization errors: anomaly-free areas are misjudged as defects. We assume that FLM can serve as an effective complement to ILM for achieving finer-grained construction. To verify this hypothesis, we visualize some features extracted by MRKD from a typical category, hazelnut. Specifically, we perform principal component analysis (PCA) on the first layer of the student network to extract the principal features with a channel of three. As shown in Fig. 9, there are some distortions (marked with red rectangles) in the features yielded by MRKD *without FLM*. By contrast, the MRKD *with FLM* alleviates distortions, obtaining more accurate features (marked with red rectangles). Moreover, we exhibit heatmaps of localization results. As shown in Fig. 10, the introduction of FLM significantly mitigates localization errors and promotes the precision of anomaly localization, which further highlights its necessity.

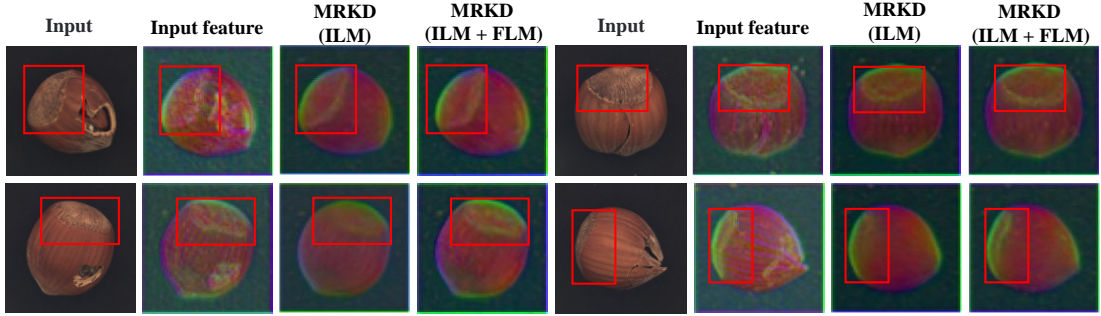

**Fig. 9.** Visualization, using PCA, of the features generated by MRKD. **Input:** Test images with defects. **Test feature:** Input features of the student network. **MRKD (ILM):** Restored features generated by the student network in MRKD considering only ILM. **MRKD (ILM + FLM):** Restored features generated by the student network in MRKD considering both ILM and FLM.

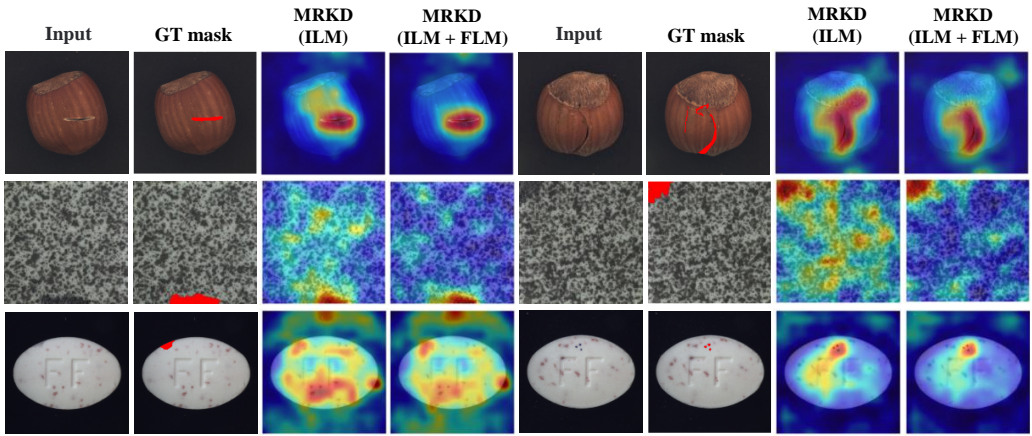

**Fig. 10.** Anomaly localization results. **Input:** Test images with defects. **GT mask:** Input with ground-truth localization regions masked in red. **MRKD (ILM):** Anomaly localization heatmaps produced by the student network in MRKD considering only ILM. **MRKD (ILM + FLM):** Anomaly localization heatmaps produced by the student network in MRKD considering both ILM and FLM.

**Contribution of the masking proportion.** We further investigate the impact of the masking ratio $\alpha$ in ILM: the proportion of synthetic abnormal samples in the training set is defined as $\alpha$. As shown in Table 6, when the value of $\alpha$ is 0, the network exhibits outstanding anomaly detection performance but unsatisfactory localization ability. This can be attributed to the fact that the network can easily restore features with only a few pixels distinguishing from input, but it struggles to restore discriminative features that significantly deviate from the entire abnormal regions. Then, as $\alpha$ rises from 0 to 0.2, the network's ability to localize anomalies improves, resulting in increasing AU-ROC$_{AL}$ and AU-PRO scores. However, as $\alpha$ continues to increase, this growth trend becomes less apparent. Our findings suggest that even a small number of synthetic abnormal samples can provide valuable information to the model. However, as the number of abnormal samples increases, the model may struggle to find additional unexplored knowledge, resulting in a slower growth rate of localization accuracy. In conclusion, localization capability can be significantly enhanced by learning only a few synthetic abnormal samples.

**Table 6**

Ablation study on masking ratio $\alpha$ in ILM. $\alpha$ represents the proportion of synthesis abnormal pictures in the training pictures. The best results are in bold.

| $\alpha$ | $\alpha = 0$ | $\alpha = 0.2$ | $\alpha = 0.4$ | $\alpha = 0.6$ | $\alpha = 0.8$ | $\alpha = 1$ |
|---|---|---|---|---|---|---|
| AU-ROC$_{IL}$ | 98.9 | 98.9 | 98.9 | 98.9 | 98.9 | **98.9** |
| AU-ROC$_{PL}$ | 98.0 | 98.3 | 98.3 | 98.3 | 98.3 | **98.4** |
| AU-PRO | 94.4 | 95.0 | 95.2 | 95.1 | 95.1 | **95.3** |

We also illustrate the influence of the masking ratio $\lambda$ in FLM on anomaly detection and localization. As shown in Fig. 11, the performance reaches its peak when $\lambda$ equals 0.2. However, as the $\lambda$ increases beyond this point, the overall performance of the network decreases.
Note that when $\lambda$ is up to 0.8, there is an intensified degradation. This can be ascribed to the fact that a very high portion of information loss biases the network towards learning global information instead of local information to achieve restoration, thus losing fine-grained features.

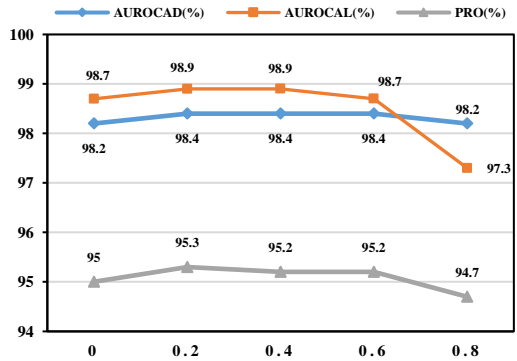

**Fig. 11.** The influence of the masking ratio $\lambda$ in FLM. X-axis: the value of $\lambda$.

**Contribution of multi-scale feature fusion.** Experiments are carried out to analyze the influence of the multi-scale feature in discriminability for anomaly detection and localization. As shown in Table 7, $SM^2$ obtains the best result among all single-layer features due to a medium receptive field capable of comprehensively extracting both textural and structural information. Furthermore, pairwise feature fusion and three-layer feature fusion all outperform the single-layer feature, which validates the necessity of feature fusion.

To intuitively illustrate the properties of different features, we select four categories with two defect types: structure and texture defects, then visualize the performance of anomaly localization in Fig. 12. In samples from the transistor and bottle categories with large-scale structural defects, the fusion of multi-scale features in the deep features outperforms shallow features in anomaly localization. This is because shallow features contain mainly low-scale textural information, while deeper features excel at extracting large-scale structure information conducive to localizing structural anomalies. In addition, the anomalies of the selected hazelnut and wood categories are classified as textural defects. Multi-scale feature fusion also demonstrates superiority over single-layer features on such defects. To be specific, $SM^1$ can localize these textural anomalies, while suffering from misdiagnosing some normal textures as defects owning to the ignoring of global structure. $SM^3$ and fusion feature map $SM^{2\&3}$ mitigate the misdiagnosis, while having difficulty in pinpointing anomalies. Therefore, the fuse of shallow and deep features can combine the information of both texture and structure, thus gaining the accuracy of anomaly localization.

**Table 7**

Ablation study on multi-scale feature fusion. The best results are in bold.

| Score Map | $SM^1$ | $SM^2$ | $SM^3$ | $SM^{1\&2}$ | $SM^{2\&3}$ | $SM^{1\&2\&3}$ |
|---|---|---|---|---|---|---|
| **AU-ROC$_{IL}$** | 94.2 | 98.0 | 97.3 | 98.1 | 98.3 | **98.9** |
| **AU-ROC$_{PL}$** | 95.4 | 97.9 | 97.2 | 97.8 | 98.2 | **98.4** |
| **AU-PRO** | 91.5 | 94.4 | 84.3 | 94.7 | 94.3 | **95.3** |

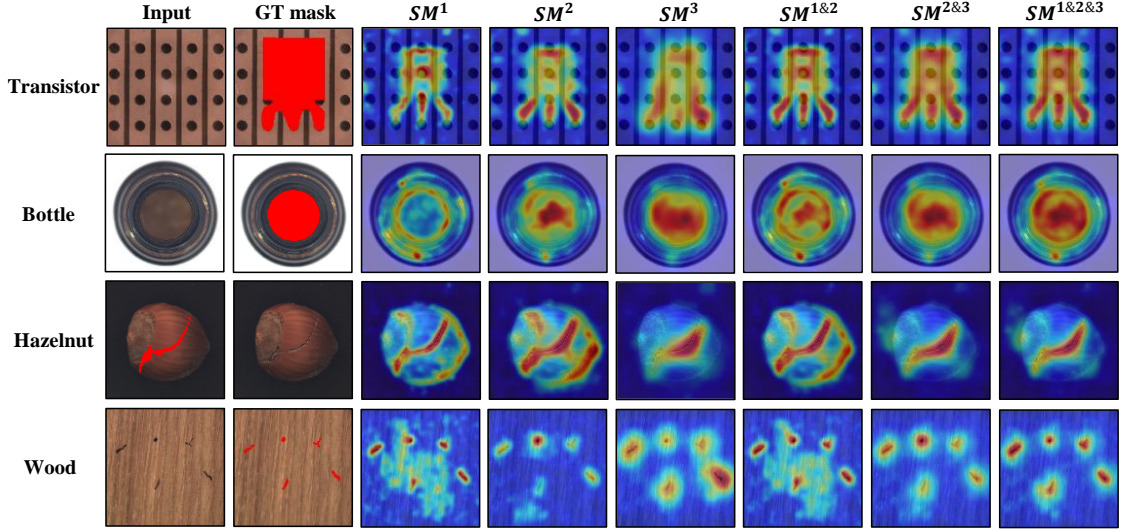

**Fig. 12.** The influence of the multi-scale feature fusion. **Input:** Test images with defects. **GT mask:** Input with ground-truth localization regions masked in red. $SM^1, SM^2, SM^3$ : Anomaly localization maps yielded by features from the first to the third layer, respectively. $SM^{1\&2}, SM^{2\&3}, SM^{1\&2\&3}$: Anomaly localization heatmaps yielded by multi-scale feature fusion.

**Contribution of backbones.** To verify the effectiveness of MRKD, different backbones (ResNet18, ResNet50, and WideResNet50) are evaluated. Table 8 indicates that as the backbone deepens and widens, the capability of the network to identify anomalies is improved. This is because ResNet50 and WResNet50 with considerably increased depth can extract high-level semantic feature that promotes the discriminability of the proposed MRKD. Moreover, the study found that even a small network like ResNet18 with fewer blocks can achieve satisfactory anomaly detection and location results using MRKD despite its limited knowledge distillation capability.

**Table 8**

Ablation study on different backbones. The best results are in bold.

| Backbones | ResNet18 | ResNet50 | WResNet50 |
|---|---|---|---|
| **AU-ROC$_{IL}$** | 98.0 | 98.7 | **98.9** |
| **AU-ROC$_{PL}$** | 97.8 | 98.3 | **98.4** |
| **AU-PRO** | 94.0 | 95.0 | **95.3** |

**Limitations.** We have observed certain limitations of the approach to detecting anomalies in the toothbrush and zipper datasets. Specifically, we have identified a decrease in performance attributed to the misclassification of impurities, such as dust particles, as anomalies. Notably, the presence of impurities, such as gray scratches on the toothbrush background or white dust on a zipper, has led to the misclassification of normal samples as anomalies, as shown in Fig. 13. To address this issue, it is crucial to develop a comprehensive understanding of the intrinsic features exhibited by genuine anomalies. Authentic anomalies possess distinct semantic characteristics that differentiate them from normal samples, as opposed to purely relying on visual appearances. Therefore, further research should focus on exploring semantic cues and incorporating them into our method to improve the discrimination between true anomalies and common impurities.

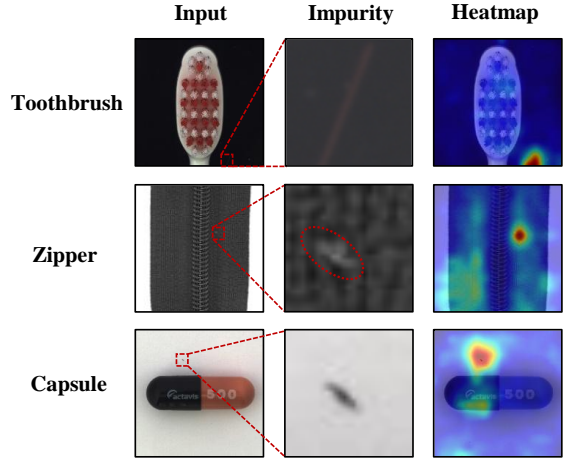

Fig. 13. Visualization of impurities. **Input:** Normal images in test data with some tiny impurities. **Impurity:** Enlarged view of the impurity areas. In order to enhance the visibility of impurities in the toothbrush class, the transparency of the image has been adjusted. **Heatmap:** Anomaly localization heatmaps of MRKD.

## 5. Conclusions and Discussions

This paper proposes a method called Masked Reverse Knowledge Distillation (MRKD), combining image-level masking (ILM) and feature-level masking (FLM) strategies to alleviate the overgeneralization problem. Essentially, MRKD takes full account of both global and local information to restore anomalies to normal appearance with the assistance of different synthetic anomalies in ILM and FLM. In ILM, the synthetic anomalies are regarded as semantic entities rather than pixels, thereby encouraging the student network to be biased more toward global information. Meanwhile, in FLM, features yielded by the student network are randomly masked for synthesizing feature-level anomalies. The restoration process of these anomalies further enriches the local information within the learned representations. Combining these two designs yields the high discriminability AD model: MRKD has shown a state-of-the-art performance on three real-world anomaly detection datasets. Moreover, ablation studies demonstrate that the proposed MRKD can produce accurate normal features recovered from anomalies, which validates its effectiveness in mitigating the overgeneralization problem.

In the future, additional research could be done to further deepen hidden representations inside the network for learning the semantics of normal and abnormal data. To fulfill this purpose, improving model architectures and introducing new data augmentations are promising directions. For instance, we can augment CNN with a transformer model to capture long-range dependencies or distort the shape of input images to bias the network more towards structural information. Furthermore, the model adaptation ability of this research can also be extended to learn image context in various anomaly categories to develop a category-agnostic anomaly detection model.

# Acknowledgments

This work has been partially supported by the Fundamental Research Funds for the Central Universities of China (2021GCRC058).